\documentclass[12pt]{article}

\usepackage{authblk} 
\usepackage{hyperref} 

\RequirePackage{amsthm,amsmath}

\RequirePackage{threeparttable}

\usepackage{amssymb,amsfonts}
\usepackage{algorithmic}
\usepackage{graphicx}
\usepackage{textcomp}
\usepackage{xcolor}
\usepackage{graphicx} 
\usepackage[authoryear]{natbib}

\usepackage{wrapfig}
\usepackage{float}

\usepackage{mathrsfs}
\usepackage{physics}
\usepackage{xcolor}
\usepackage{subcaption}
\usepackage{tabularx}

\title{Measuring Heterogeneity in Machine Learning with Distributed Energy Distance}

\author[1,*]{Mengchen Fan}
\author[1]{Baocheng Geng}
\author[2]{Roman Shterenberg}
\author[2]{Joseph A. Casey}
\author[3]{Zhong Chen}
\author[2,**]{Keren Li}

\affil[1]{Department of Computer Science, University of Alabama at Birmingham}
\affil[2]{Department of Mathematics, University of Alabama at Birmingham}
\affil[3]{School of Computing, Southern Illinois University}
\affil[*]{\textbf{First Author:} Contributed most significantly to this work.}
\affil[**]{\textbf{Corresponding Author:} \href{mailto:ki@uab.edu}{ki@uab.edu}}

\date{} 

\begin{document}

\maketitle

\begin{abstract}
In distributed and federated learning, heterogeneity across data sources remains a major obstacle to effective model aggregation and convergence. We focus on feature heterogeneity and introduce energy distance as a sensitive measure for quantifying distributional discrepancies. While we show that energy distance is robust for detecting data distribution shifts, its direct use in large-scale systems can be prohibitively expensive. To address this, we develop Taylor approximations that preserve key theoretical quantitative properties while reducing computational overhead. Through simulation studies, we show how accurately capturing feature discrepancies boosts convergence in distributed learning. Finally, we propose a novel application of energy distance to assign penalty weights for aligning predictions across heterogeneous nodes, ultimately enhancing coordination in federated and distributed settings.
\end{abstract}



\section{Introduction}\label{sec:intro}

Distributed and federated learning systems have emerged as powerful paradigms for training large-scale machine learning models across geographically dispersed data sources. By coordinating learning processes among multiple nodes (clients) without centrally aggregating raw data, these systems address privacy and communication constraints. However, a persistent and critical challenge in these settings is \textbf{heterogeneity}---the variations in data, resources, tasks, or network characteristics across different nodes---that can significantly degrade both model accuracy and convergence performance. As real-world federated learning (FL) deployments diverge from the ideal assumption of identical and independently distributed (IID) data  \cite{li2023nn,geng2021collaborative}, understanding and mitigating heterogeneity has become a focal point of research.

Several forms of heterogeneity have been identified in the literature. \textit{Task heterogeneity} arises when nodes optimize distinct learning objectives or adopt different model architectures, complicating global model aggregation unless task-specific adaptations are employed \cite{chen2021hetmaml,shen2023episodic}. \textit{Network heterogeneity} involves disparities in communication bandwidth, latency, or reliability among nodes, creating asynchronous updates and delayed convergence \cite{gures2022machine}. Similarly, \textit{resource heterogeneity}, reflecting differences in computational power, memory, or energy availability, can lead to imbalanced participation and hinder overall system performance \cite{tyagi2020taming}. Among these variations, however, \textbf{data heterogeneity} (i.e., non-IID data distributions across nodes) is arguably the most pervasive and, in many cases, the most detrimental to learning performance \cite{kamm2023survey}.
Within data heterogeneity, researchers have highlighted several specific scenarios that can arise:
\begin{enumerate}
    \item \textbf{Model heterogeneity} (also called conditional distribution heterogeneity) focuses on situations where the marginal distribution of features, \(P(X)\), remains consistent, but the conditional distribution, \(P(Y \mid X)\), diverges across nodes. Methods such as personalized federated learning, multi-task learning, or domain adaptation have been proposed to bridge these discrepancies by tailoring global models to local distributions while retaining shared knowledge \cite{lin2020ensemble,alam2022fedrolex}.

    \item \textbf{Label heterogeneity} refers to variations in the marginal distribution of labels, \(P(Y)\), while the conditional feature distribution, \(P(X \mid Y)\), stays consistent. Such shifts in label priors affect class imbalance and can cause biases in global model aggregation. Techniques like FedProx \cite{li2020federated} and Ditto \cite{li2021ditto} seek to regularize local updates toward a global objective, mitigating the adverse effects of unbalanced labels.

    \item \textbf{Feature heterogeneity} arises when the marginal distribution \(P(X)\) differs across nodes, even while the conditional \(P(Y \mid X)\) remains stable. This can occur due to different data collection protocols, sensor types, or preprocessing methods, posing significant challenges for representation-learning models. Domain generalization and data augmentation \cite{hao2021towards,xu2023bias,zhang2023data}, as well as client clustering \cite{yan2023clustered}, have been explored to address feature shifts, but the topic remains relatively under-studied compared to label heterogeneity.

    \item \textbf{Representation heterogeneity} captures scenarios where \(P(Y)\) remains consistent, but \(P(X \mid Y)\) diverges. Nodes may learn different internal representations for the same label, complicating model transferability and aggregation. Approaches such as contrastive learning \cite{zheng2022contrastive}, domain adaptation \cite{liu2020heterogeneous} and representation learning \cite{10706324} aim to align these representations but often incur considerable computational or communication overhead.
\end{enumerate}

Despite the abundance of methods proposed to tackle various forms of heterogeneity, a comprehensive, quantitative measure of heterogeneity, and a rigorous analysis of its influence on model performance, remains an open challenge. In particular, while metrics like KL divergence can capture certain distributional differences, they may fail to detect more subtle shifts in feature or representation spaces. This motivates the need for a more sensitive and robust approach to quantifying distributional divergences in high-dimensional settings.

\textbf{Our Work and Contributions:}
In this paper, we focus on \textbf{feature heterogeneity} and present a novel framework for quantitatively assessing and mitigating its impact on distributed learning systems. Specifically, we propose the use of the \textit{energy distance} to measure discrepancies between feature distributions across different nodes. Unlike some conventional divergence measures, the energy distance is sensitive to both location and scale differences and maintains robustness in high-dimensional spaces. However, directly applying the energy distance in large-scale systems can be computationally expensive. To address this limitation, we introduce a \textbf{Taylor approximation} that retains the key properties of the energy distance while ensuring practical scalability.

Our key contributions include:
\begin{itemize}
    \item \textbf{Heterogeneity Measurement}: We introduce the energy distance metric as a principled way to quantify the degree of feature heterogeneity across nodes, offering finer insights into the distributional shifts that degrade model performance.
    \item \textbf{Efficient Approximation}: We develop a Taylor approximation approach that significantly reduces the computational overhead of the energy distance, making it suitable for large-scale, distributed environments.
    \item \textbf{Empirical Evaluation}: Through extensive experiments, we demonstrate how this heterogeneity measure correlates with learning performance and propose strategies to adapt global models accordingly in both federated and broader distributed learning scenarios.
\end{itemize}

Our findings highlight the critical role of accurately assessing feature heterogeneity and pave the way for more robust distributed learning systems that can adapt to real-world, non-IID data distributions. We believe this line of research will foster deeper insights into heterogeneity's consequences and lead to improved algorithmic strategies in federated and distributed machine learning.

\section{Distributed Energy Distance for Feature Heterogeneity}\label{sec:metric_hetero}

Understanding and quantifying \textbf{feature heterogeneity} is pivotal in distributed learning systems, as it directly impacts model performance and convergence. Feature heterogeneity arises when the marginal distributions of predictors differ across nodes, leading to challenges in federated or distributed algorithms that assume data homogeneity. A robust metric is required to characterize such heterogeneity and facilitate hypothesis testing to assess whether features at different nodes are drawn from the same underlying distribution.

To address this need, we leverage the \textbf{energy distance}, a versatile metric that quantifies distributional differences by comparing pairwise distances between samples while accounting for intra-distribution variability. This measure highlights significant disparities between feature distributions and forms the foundation for adaptive strategies to address heterogeneity-induced challenges in distributed systems.

This framework sets the stage for exploring energy distance in distributed scenarios, where constraints on communication and computational resources necessitate efficient approximations and tailored hypothesis testing methods.

The literature offers a variety of metrics to quantify distances between datasets, ranging from classical statistical measures to modern machine-learning-inspired approaches. The choice of metric plays a critical role in accurately capturing distributional differences and addressing the challenges posed by feature heterogeneity in distributed systems.

\subsection{Energy Distance and Energy Coefficient}\label{sec:energy}

The \textbf{Energy Distance}, introduced by \cite{szekely2003statistics, szekely2005new}, is a non-parametric measure that quantifies differences between distributions by evaluating pairwise distances between samples, while accounting for the internal variability within each distribution. For two random vectors $X$ and $Y$, the squared energy distance is defined as
\begin{align*}
 D^2(X, Y) = 2 \mathbb{E} \norm{X - Y} - \mathbb{E} \norm{X - X'} - \mathbb{E} \norm{Y - Y'},
\end{align*}
where $\norm{\cdot}$ represents the $L_2$ norm, and $X', Y'$ are independent copies of $X, Y$ respectively. Its empirical version, the Energy statistic, calculated from datasets $\mathbf{x}=\{x_i\}_{i=1}^n$ and $\mathbf{y}=\{y_j\}_{j=1}^m$, is given by
\begin{align}\label{eq:emp}
 E_{n, m}(\mathbf{x}, \mathbf{y}) = & 2 \cdot \frac{1}{nm} \sum_{i=1}^n \sum_{j=1}^m \norm{x_i - y_j} \notag \\
 & - \frac{1}{n^2} \sum_{i=1}^n \sum_{j=1}^n \norm{x_i - x_j} - \frac{1}{m^2} \sum_{i=1}^m \sum_{j=1}^m \norm{y_i - y_j} 
\end{align}

The \textbf{Energy Coefficient} $H$ extends the energy distance into a normalized measure of heterogeneity
\begin{align*}
 H = \frac{D^2(X, Y)}{2 \mathbb{E} \norm{X - Y}},
\end{align*}
where $H \in [0, 1]$, with $H = 0$ if and only if $P_X = P_Y$. This coefficient captures general distributional differences, including discrepancies in location, scale, and shape, while offering an interpretable measure suitable for comparisons across datasets.

The Energy Distance possesses several desirable properties, making it a robust and versatile metric. It satisfies the fundamental requirements of a distance metric: non-negativity, symmetry, and the triangle inequality. Furthermore, it is invariant to scaling, rotation, and translation, ensuring consistent behavior under affine transformations. The Energy Distance is also sensitive to differences in location, scale, and higher-order moments, making it particularly well-suited for comparing distributions in both low-dimensional and high-dimensional settings. These properties, combined with its asymptotic unbiasedness, make the Energy Distance a compelling choice for analyzing distributional differences in various scenarios.

This framework inherently supports hypothesis testing. For two datasets $\mathbf{x}$ and $\mathbf{y}$, with observed pairwise distance matrices $D(X,Y)$, $D(X,X)$, and $D(Y,Y)$, energy distance enables evaluation of the null hypothesis $H_0: P_X = P_Y$. The test statistic derived from the energy distance asymptotically follows a known distribution under $H_0$, allowing for direct computation of $p$-values without requiring computationally intensive methods like permutation tests.

Energy Distance stands out among metrics for measuring distributional differences due to its computational simplicity and broad applicability. Unlike \textbf{Maximum Mean Discrepancy (MMD)}, which relies on kernel selection in reproducing kernel Hilbert spaces, or \textbf{Wasserstein Distance}, which is computationally intensive, Energy Distance avoids these complexities while capturing both location and scale differences. Although metrics like \textbf{Kullback-Leibler (KL) Divergence} and \textbf{Total Variation Distance} are useful in specific scenarios, they struggle with non-overlapping supports or high-dimensional data. Overall, Energy Distance provides a robust and versatile framework for analyzing heterogeneous datasets, making it particularly effective for high-dimensional distributed learning.

\subsection{Comparison with other metrics}

In the landscape of metrics for measuring distributional differences, several alternatives to Energy Distance are widely used, each with distinct advantages and limitations.

\textbf{Maximum Mean Discrepancy (MMD)} works by mapping distributions into a reproducing kernel Hilbert space (RKHS) and measuring their distance. Its performance depends on the choice of kernel, which can be difficult to tune, especially in high-dimensional settings.

\textbf{Wasserstein Distance}, or Optimal Transport, calculates the cost of transforming one distribution into another, making it ideal for capturing geometric differences. However, it is computationally intensive, particularly in high dimensions.

\textbf{Kullback-Leibler (KL) Divergence} measures how one probability distribution diverges from another. While effective for comparing probability densities, it fails when the distributions lack overlapping supports and is not symmetric.

\textbf{Total Variation Distance} assesses the maximum difference in probabilities across distributions. It is straightforward for discrete cases but becomes less practical for continuous or high-dimensional data.

Energy Distance distinguishes itself by being computationally simple and broadly applicable. It avoids reliance on kernels or optimization and captures both location and scale differences. These properties make it a robust choice for analyzing heterogeneous datasets, particularly in high-dimensional scenarios.

The choice of metric depends on the task. For geometric insights, Wasserstein or Energy Distance works well, while kernel-based methods like MMD offer flexibility. Simpler alternatives like Total Variation or KL Divergence may suffice for certain cases but lack the generality and robustness of Energy Distance.

\subsection{Efficient Approximation of Distributed Energy Distance}

Energy Distance effectively quantifies distributional differences by comparing pairwise distances between samples. However, its quadratic computational complexity with respect to the number of points in $X$ and $Y$, due to the need to compute all pairwise distances, which makes it infeasible for large-scale distributed datasets, where transmitting raw data is impractical. To address this, we propose an approximation framework that uses summary statistics such as means, variances, and higher-order moments, reducing complexity to linear in sample size. \textit{In this paper, we present only the final results of the efficient approximation due to space constraints. Detailed derivations will be provided in the extended version of this manuscript.}

For one-dimensional cases, let $X$ and $X'$ be independent and identically distributed random variables with mean $\mu$, variance $\sigma^2 $, skewness $\gamma_3$, and excess kurtosis $\gamma_4$. 

First, consider the squared distance between $X$ and $X'$,
\begin{align*}
    \mathbb{E}\left[\norm{X - X'}^2\right] = 2 \sigma^2.
\end{align*}

We are interested in approximating the expectation of the square root of this distance, which is given by:
\begin{align*}
\mathbb{E}\left[\norm{X - X'}\right] = \mathbb{E}\left[\sqrt{(X - X')^2}\right],
\end{align*}
with $\sqrt{2} \sigma$ its upper bound by Jensen's Inequality.

A Taylor series expansion is applied around the mean of $(X - X')^2$, which is $2 \sigma^2$. Let $g(z) = \sqrt{z}$. The  second-order Taylor expansion of $g(z)$ around $z = 2 \sigma^2$ is given by
\begin{align*}
g(z) &\approx g(2 \sigma^2) + g'(2 \sigma^2) \left(z - 2 \sigma^2\right) + \frac{1}{2} g''(2 \sigma^2) \left(z - 2 \sigma^2\right)^2\\
&=\sqrt{2 \sigma^2} + \frac{(X - X')^2 - 2 \sigma^2}{2 \sqrt{2 \sigma^2}} - \frac{\left((X - X')^2 - 2 \sigma^2\right)^2}{8 (2 \sigma^2)^{3/2}}.
\end{align*}

After expectation, the second term on the right-hand side vanishes, and the third term involves the fourth central moment of $X$. It leads to the final approximation
\begin{align}\label{eq:approx_xx}
\mathbb{E}\left[\norm{X - X'}\right] \approx \sqrt{2} \sigma \left(1 - \frac{\gamma_4 +4}{16}\right).
\end{align}
The correction term $- \frac{\sqrt{2} \gamma_4 }{16} \sigma$ accounts for the effect of kurtosis on the expected distance.

For two independent random variables $X$ and $Y$ with means $\mu_X $ and $\mu_Y$, variances $\sigma_X^2$ and $\sigma_Y^2$, skewnesses $\gamma_{3X}$ and $\gamma_{3Y}$, and kurtoses $\gamma_{4X}$ and $\gamma_{4Y}$. Similarly, the expectation of the Euclidean distance between $X$ and $Y$ can be approximated by
\begin{small}
\begin{align}\label{eq:approx_xy}
 \mathbb{E} \left[ \norm{X - Y} \right] \approx \sqrt{\nu_{XY}} \left(1 - \frac{C_{4XY} + 4 C_{3XY} \delta_\mu + 2\nu_{XY}^2 - 2\delta_\mu^4}{8 \nu_{XY}^2}\right),
\end{align}
\end{small}
where $\nu_{XY} = \sigma_X^2 + \sigma_Y^2 + \delta_\mu^2$, $C_{4XY} = \sigma_X^4 \gamma_{4X} + \sigma_Y^4 \gamma_{4Y}$, and $C_{3XY} = \sigma_X^3 \gamma_{3X} - \sigma_Y^3 \gamma_{3Y}$, $\delta_\mu=\mu_X-\mu_Y$.

The leading term $\sqrt{\nu_{XY}}$ captures the total dispersion between $X$ and $Y$, while the correction term $\frac{C_{4XY} + 4 C_{3XY} \delta_\mu + 2\nu_{XY}^2 - 2\delta_\mu^4}{8 \nu_{XY}^2}$ adjusts for the effects of skewness and kurtosis in the distributions of $X$ and $Y$. This approximation is consistent with the approximation \eqref{eq:approx_xx} when $X\sim Y$.

For the $d$-dimensional case, let $X = (X_1, X_2, \dots, X_d)$ and $Y = (Y_1, Y_2, \dots, Y_d)$ be independent random vectors with component means $\mu_{X_i} = \mathbb{E}[X_i]$ and $\mu_{Y_i} = \mathbb{E}[Y_i]$, variances $\sigma_{X_i}^2 = \text{Var}(X_i)$ and $\sigma_{Y_i}^2 = \text{Var}(Y_i)$, skewnesses $\gamma_{3X_i}$ and $\gamma_{3Y_i}$, and kurtoses $\gamma_{4X_i}$ and $\gamma_{4Y_i}$ for $i = 1, \dots, d$.

The expectation of the Euclidean distance between $X$ and $X'$ is approximated as
\begin{align*}
 \mathbb{E} \left[ \norm{X - X'} \right] \approx \sqrt{\nu_{XX'}} \left(\frac{3}{4} - \frac{C_{4X}}{8 \nu_{XX'}^2}\right),
\end{align*}
where $\nu_{XX'} = 2\sum_{i=1}^d \sigma_{X_i}^2$, $C_{4X} = 2\sum_{i=1}^d \sigma_{X_i}^4 \gamma_{4X_i}$.

The expectation of the Euclidean distance between $X$ and $Y$ is approximated by
\begin{align}\label{eq:approx_xyd}
 \mathbb{E} \left[ \norm{X - Y} \right] \approx \sqrt{\nu_{XY}} \left(\frac{3}{4} - \frac{C_{4XY} + 4 C_{3XY}  \delta_{\mu 1} - 2 (\delta_{\mu 2})^2}{8 \nu_{XY}^2}\right),
\end{align}
where $\nu_{XY} = \sum_{i=1}^d \left( \sigma_{X_i}^2 + \sigma_{Y_i}^2 + (\mu_{X_i} - \mu_{Y_i})^2 \right)$, 
 $C_{4XY} = \sum_{i=1}^d \left( \sigma_{X_i}^4 \gamma_{4X_i} + \sigma_{Y_i}^4 \gamma_{4Y_i} \right)$, 
 $C_{3XY} = \sum_{i=1}^d \left( \sigma_{X_i}^3 \gamma_{3X_i} - \sigma_{Y_i}^3 \gamma_{3Y_i} \right)$, 
 $\delta_{\mu 1} = \sum_{i=1}^d (\mu_{X_i} - \mu_{Y_i})$, and 
 $\delta_{\mu 2} = \sum_{i=1}^d (\mu_{X_i} - \mu_{Y_i})^2$.

The computational complexity of the empirical formula is $O(n^2d)$, while the Taylor approximation reduces it to $O(nd)$, making it well-suited for distributed scenarios.

This approximation framework enables the efficient computation of Energy Distance in distributed environments, ensuring scalability and compatibility with privacy-preserving constraints.

\subsection{Residual Discrepancy and Variance Analysis}\label{sec:theory}

The Taylor approximation of $\mathbb{E}\left[\norm{X - X'}\right]$ incurs residual errors dominated by third-order terms, expressed as
\begin{align*}
\mathcal{R}_3 = \frac{3}{48 (2 \sigma^2)^{5/2}} \left(2 \kappa_6 + 18 \sigma^6 \gamma_4 + 34 \sigma^6 - 20 \sigma^6 \gamma_3^2\right).
\end{align*}
where $\kappa_6$ is the sixth cumulant of $X$, $\gamma_3$ the skewness, and $\gamma_4$ the kurtosis. These corrections are typically small but grow significant for distributions with pronounced asymmetry or heavy tails.

The variance of $\sqrt{(X - X')^2}$, approximated as
\begin{align*}
\text{Var}\left(\sqrt{(X - X')^2}\right) \approx \frac{\sigma^2 \gamma_4}{8},
\end{align*}
indicates reduced reliability for heavy-tailed distributions. However, with sufficient data aggregation, high-order effects are smoothed, minimizing practical impact despite theoretical discrepancies.

\subsection{Adjustment to Exact Expressions Using Skewness and Kurtosis}

The Taylor approximation of $\mathbb{E} \left[ \norm{X - X'} \right]$ diverges slightly from the exact expectation when $X, X' \sim N(\mu_X, \sigma_X^2)$. For Gaussian distributions, the exact expected distance is given by
\begin{align}\label{eq:exact_xx}
 \mathbb{E} \left[ \norm{X - X'} \right] &= \frac{2}{\sqrt{\pi}} \sigma.
\end{align}
This discrepancy stems from the omission of higher-order terms in the Taylor expansion.

Inspired by Hampel et al (2005) \cite{hampel2005robust}, we extend the exact formula for Gaussian distributions to handle non-Gaussian distributions by incorporating correction terms derived from skewness and kurtosis,
\begin{align}\label{eq:adj_xx}
 \mathbb{E} \left[ \norm{X - X'} \right] \approx \left(\frac{2}{\sqrt{\pi}} - \frac{\sqrt{2} \gamma_4}{16}\right) \sigma.
\end{align}
where $- \frac{\sqrt{2} \gamma_4}{16} \sigma$ accounts for excess kurtosis, improving accuracy for distributions with heavier or lighter tails.

For independent random variables $X \sim N(\mu_X, \sigma_X^2)$ and $Y \sim N(\mu_Y, \sigma_Y^2)$, the exact expectation of the Euclidean distance is
\begin{align}\label{eq:exact_xy}
 \mathbb{E} \left[ \norm{X - Y} \right] = \sqrt{\sigma_X^2 + \sigma_Y^2} \cdot \sqrt{\frac{2}{\pi}} \, e^{-\Delta^2} + \delta_\mu \left( 2 \Phi(\Delta) - 1 \right),
\end{align}
where $\Phi$ is the standard normal CDF and $\Delta = \frac{\delta_\mu}{\sqrt{2 (\sigma_X^2 + \sigma_Y^2)}}$.


Similarly, we propose the following adjustment
\begin{align}\label{eq:adj_xy}
 \mathbb{E} \left[ \norm{X - Y} \right] &\approx \sqrt{\sigma_X^2 + \sigma_Y^2} \cdot \sqrt{\frac{2}{\pi}} \, e^{-\Delta^2} + \delta_\mu \left( 2 \Phi(\Delta) - 1 \right) \notag \\
 &- (C_{4XY} + 4 C_{3XY} \delta_\mu )/(8 \nu_{XY}^\frac{3}{2}).
\end{align}

This adjustment aligns Taylor approximations with exact solutions for small $\delta_\mu$ while extending corrections for mean differences and higher moments.
It is consistent with \eqref{eq:adj_xx} for $\mathbb{E} \left[ \norm{X - X'} \right]$.


\section{Experiment Result}\label{sec:simu}

\subsection{Energy Distance Experiments}

This experiment evaluates the performance of energy distance measurements across diverse data distributions, focusing on computational efficiency and accuracy. Using a sample size of $n = 10^3$, we explored standard distributions with various degrees of skewness, kurtosis, and tail behavior, including 
Normal ($\mu=0$,$\sigma$=1, $\mu=1$,$\sigma$=1 and $\mu=10$,$\sigma=1$), Exponential ($\beta = 0.1, 1, 10$), standard Student's t ($df = 5$), Beta ($\alpha=\beta=0.5$), and Gamma ($k=1,\theta=2$). 

Two experimental setups were examined: (1) comparing samples drawn from the same type of distribution, and (2) comparing samples from each distribution to those drawn from one of the normal distributions. The computational methods assessed included the empirical energy distance formula \eqref{eq:emp}, Taylor approximation formulas \eqref{eq:approx_xx} and \eqref{eq:approx_xy}, exact formulas under normality assumptions \eqref{eq:exact_xx} and \eqref{eq:exact_xy}, and adjusted formulas incorporating skewness and kurtosis \eqref{eq:adj_xx} and \eqref{eq:adj_xy}.

\begin{figure}{H}
	\begin{center}
		\includegraphics[width=0.45\textwidth]{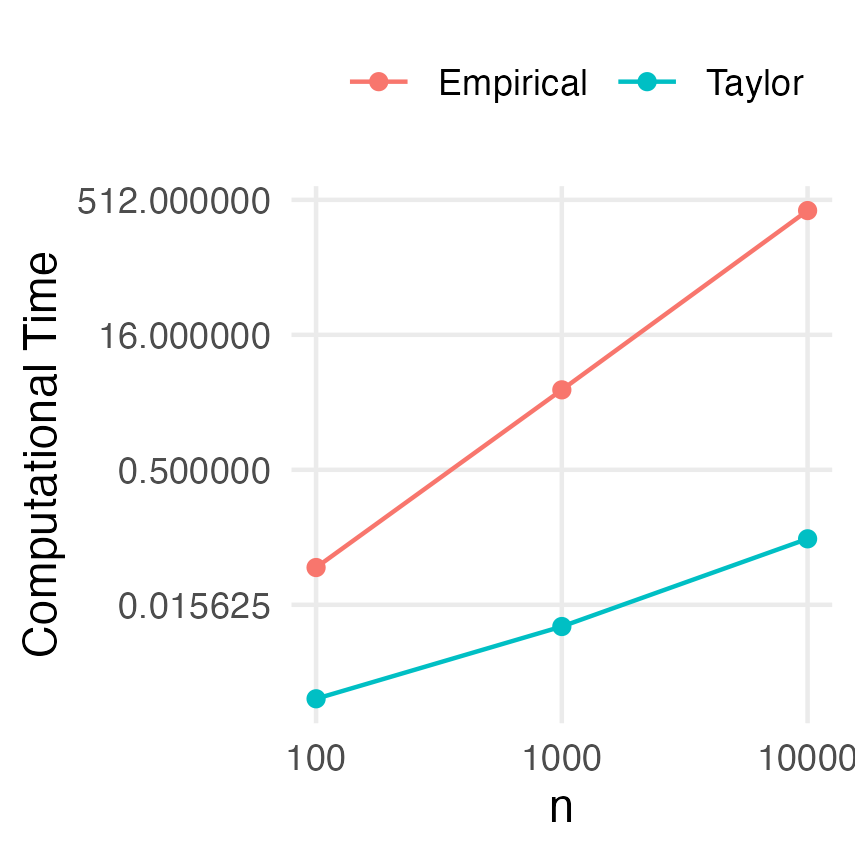}
	\end{center}
	\caption{Computational time.}
    \label{fig:ed_result2} 
\end{figure}

The Taylor approximation consistently aligned closely with empirical calculations while significantly reducing computational time, particularly as the sample size increased to $n=10^4$. As shown in Figure~\ref{fig:ed_result2}, computational time was drastically reduced using Taylor approximations, making them practical for large-scale systems. Meeting our earlier conclusions, as the sample size increased from $n = 10^2$ to $10^4$, Taylor-based methods maintained accuracy comparable to empirical calculations while significantly improving computational efficiency.

\begin{figure}[H]
 \centering


 \begin{subfigure}[b]{0.45\linewidth}
 \includegraphics[width=\linewidth]{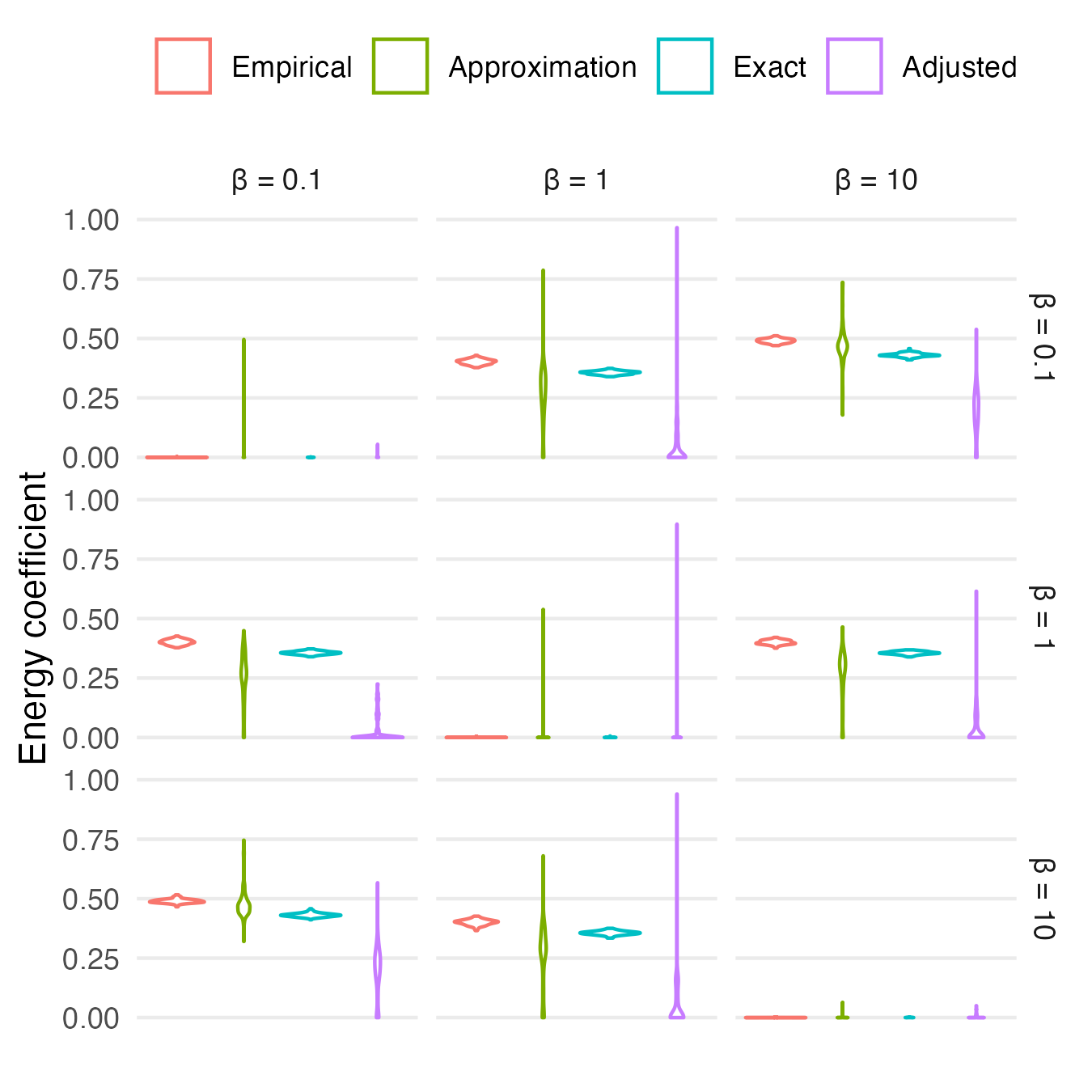} 
 \caption{Exponential} 
 \end{subfigure}
 \hfill
 \begin{subfigure}[b]{0.45\linewidth}
 \includegraphics[width=\linewidth]{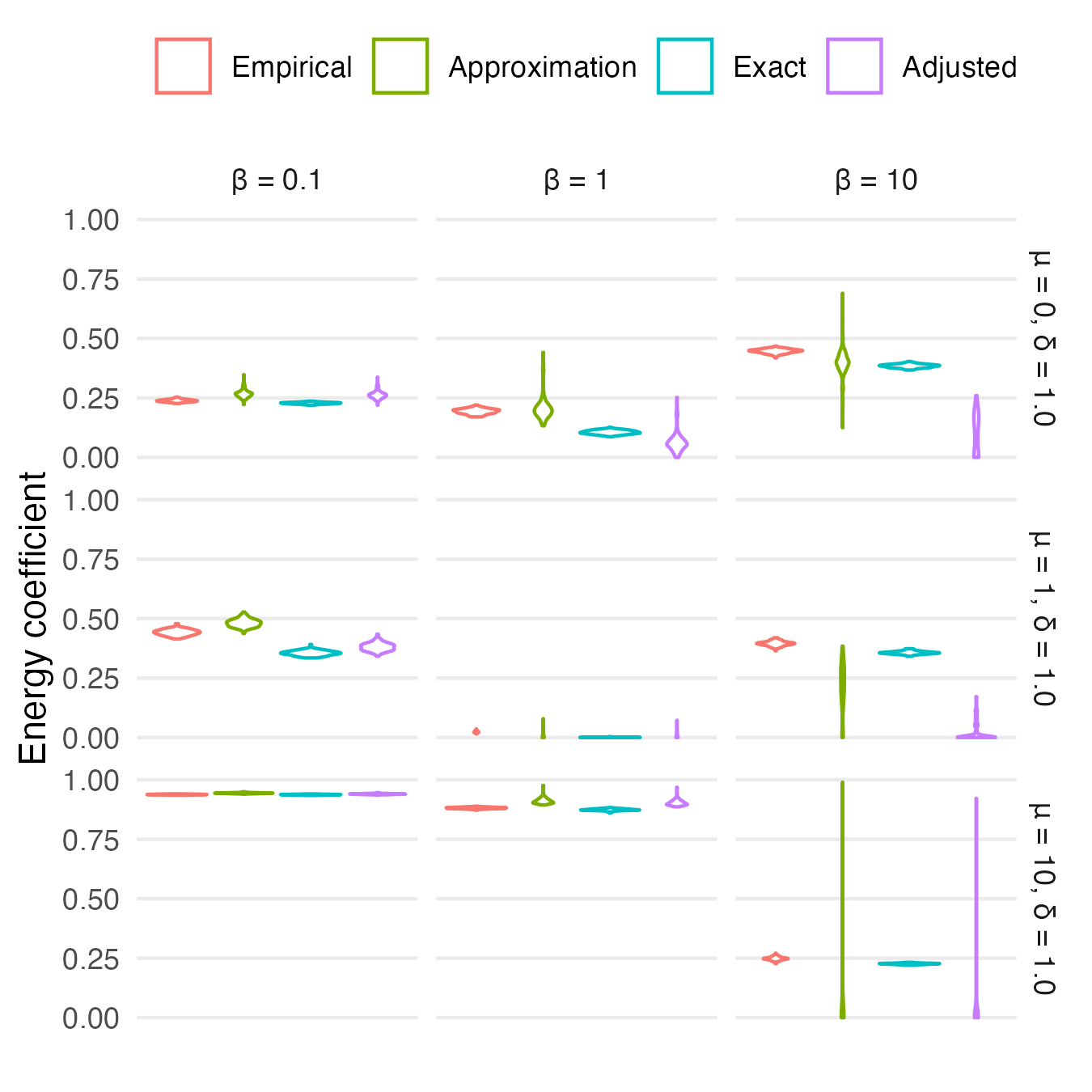} 
 \caption{Exponential vs Normal} 
 \end{subfigure}

 \begin{subfigure}[b]{0.45\linewidth}
 \includegraphics[width=\linewidth]{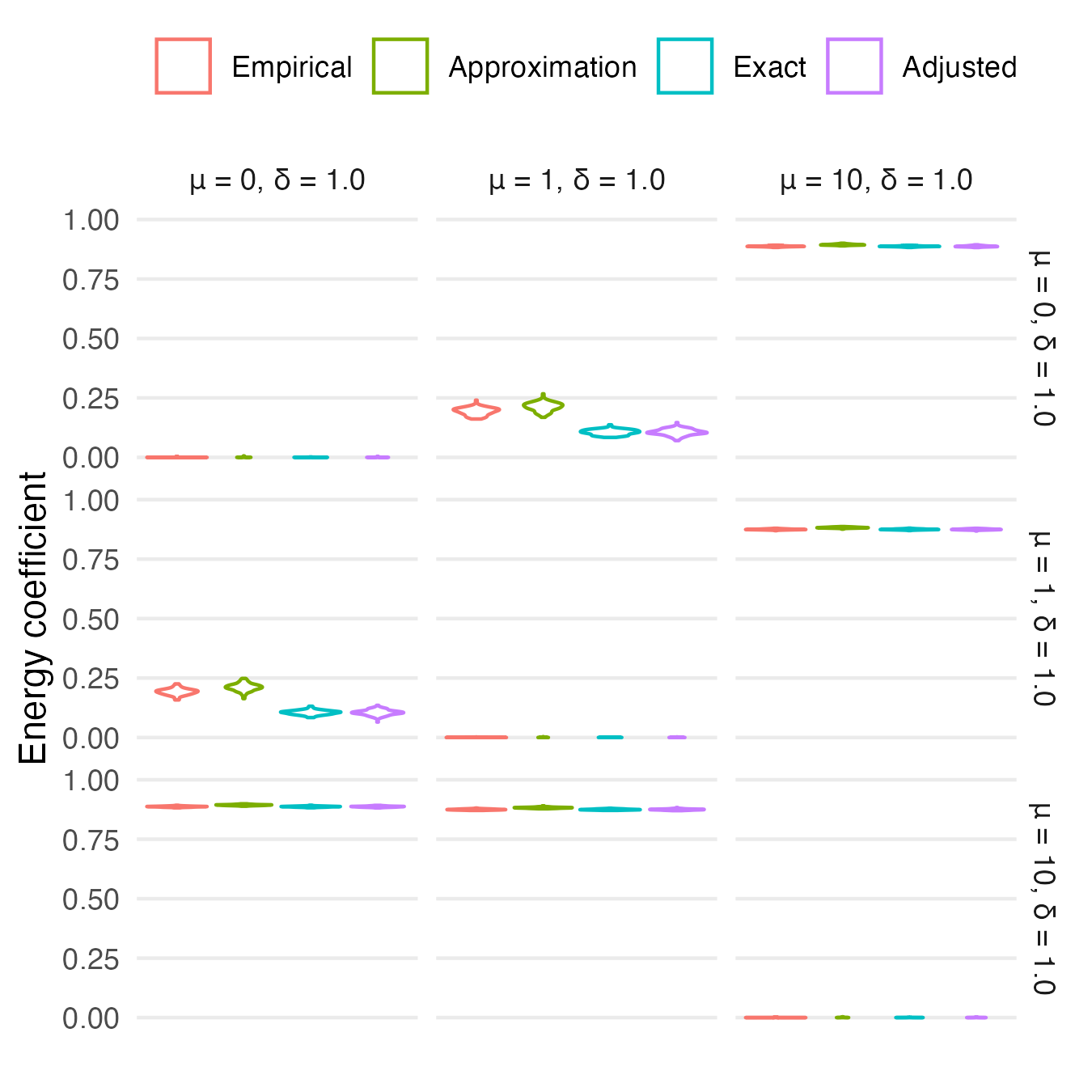} 
 \caption{Normal} 
 \end{subfigure}
 \hfill
 \begin{subfigure}[b]{0.45\linewidth}
 \includegraphics[width=\linewidth]{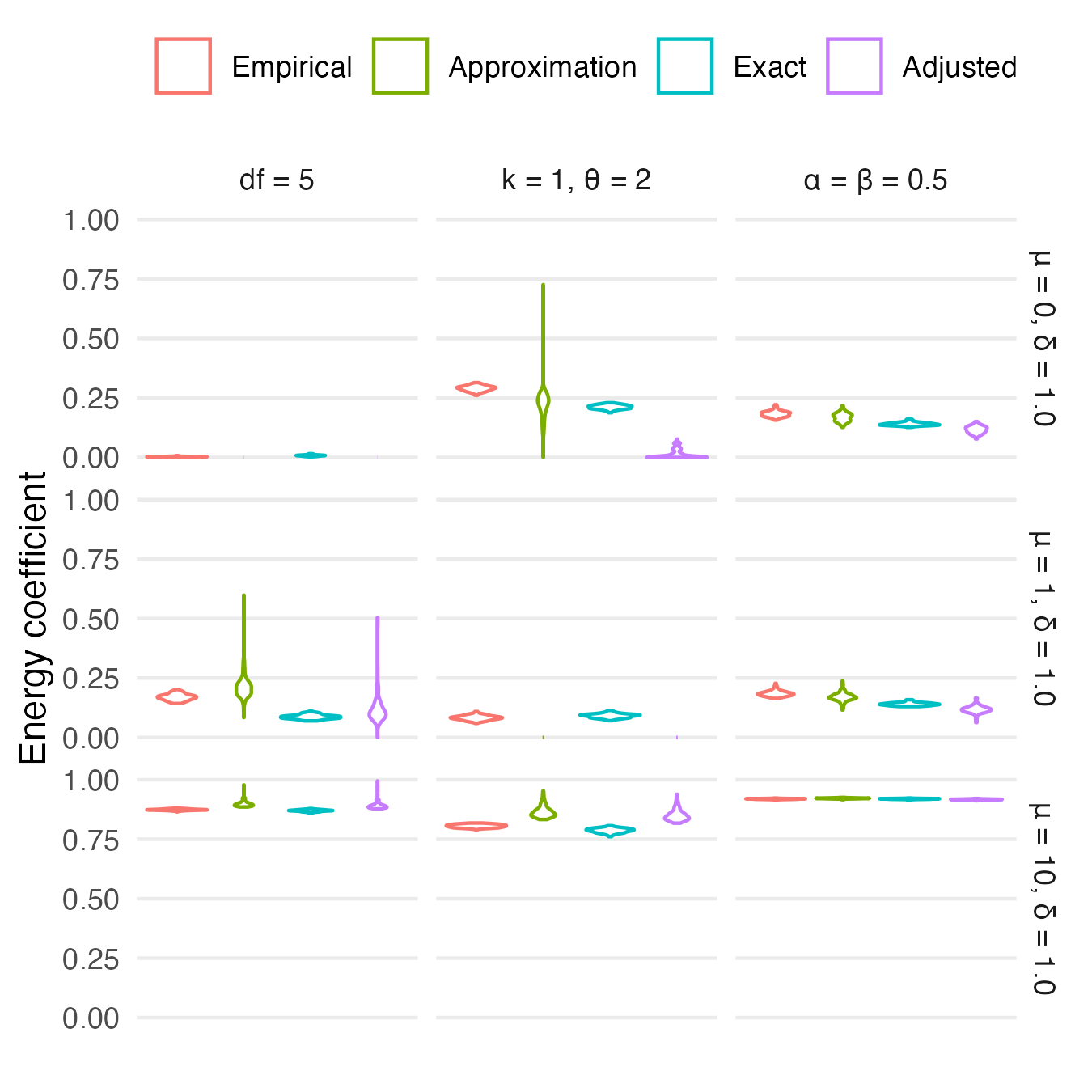} 
 \caption{Beta, Gamma, T vs Normal} 
 \end{subfigure}
 
\begin{subfigure}[b]{0.45\linewidth}
 \includegraphics[width=\linewidth]{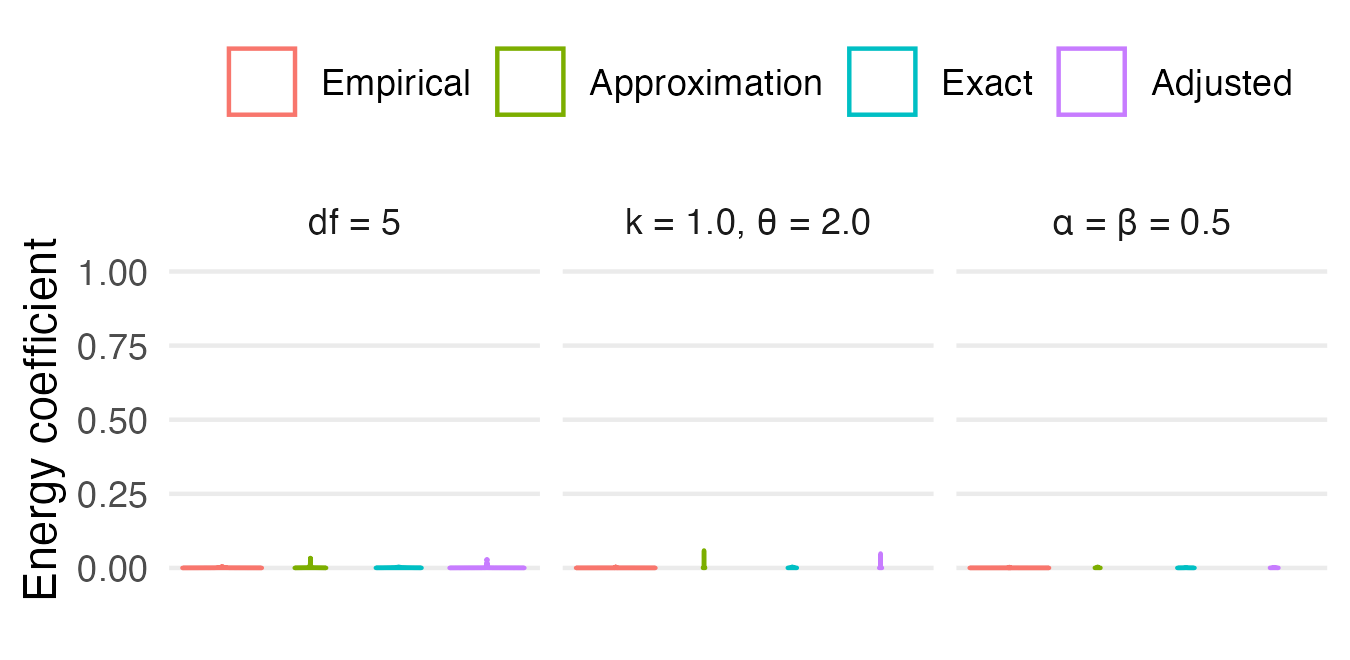} 
 \caption{Beta, Gamma, and T} 
 \end{subfigure}
 
 \caption{Energy coefficient $H$ for various distributions.}
 \label{fig:ed_result1} 
\end{figure}

Figure \ref{fig:ed_result1} summarizes the results. Across all distributions, energy coefficients $H$ computed for identical distributions reliably converged near zero, confirming the robustness of the energy distance metric for detecting distributional similarity. Even when comparing non-identical distributions, $H$ values remained low if mean differences were minimal, highlighting the dominant role of mean discrepancy in driving $H$ away from zero.

For distributions with heavy tails, such as exponential and gamma, discrepancies between computational methods were more pronounced. This emphasizes the importance of tailoring computational strategies for distributions with high skewness and kurtosis. For example, Bernoulli distributions with $p = 0.05$ ($\gamma_3 \approx 4.13$, $\gamma_4 \approx 17.68$) and $p = 0.1$ ($\gamma_3 \approx 2.67$, $\gamma_4 \approx 6.0$) demonstrated significant instability in approximation methods due to high variability in estimating higher-order moments. Similar trends were observed in exponential distributions, where $\gamma_3 = 2$ and $\gamma_4 = 6$ contributed to notable discrepancies.

These findings underscore the necessity of adapting computational methods based on the underlying distribution and scale of data. For distributions with moderate skewness and kurtosis, Taylor approximations provide an efficient and accurate alternative to empirical calculations. Conversely, exact methods may be preferred for distributions with extreme tail behavior or large mean differences. These insights establish the energy coefficient $H$ as a versatile and reliable tool for quantifying heterogeneity in distributed learning systems, paving the way for efficient energy distance estimation in large-scale applications.

\subsection{Impact of Feature Heterogeneity on Federated Learning Performance}

\begin{figure}[H]
 \centering
 \begin{subfigure}[b]{0.49\linewidth}
 \includegraphics[width=\linewidth]{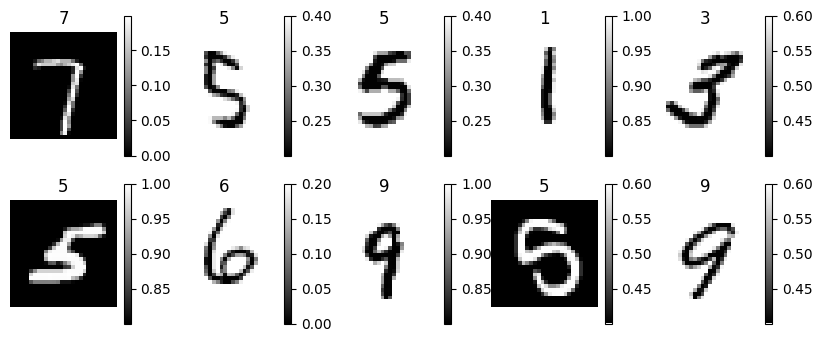} 
 \caption{Client (1)} 
 \end{subfigure}
 \hfill
 \begin{subfigure}[b]{0.49\linewidth}
 \includegraphics[width=\linewidth]{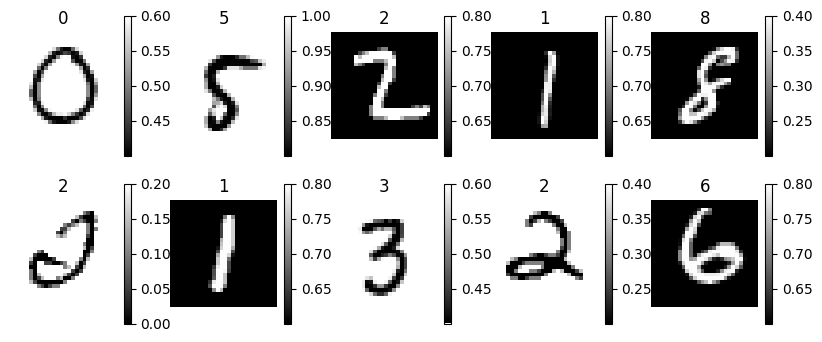} 
 \caption{Client (2)} 
 \end{subfigure}
 \caption{Mixed MNIST Inputs}
 \label{fig:fed_mix} 
\end{figure}

In this experiment, we investigate how varying feature distributions across clients impact federated learning performance. Using the MNIST dataset, a benchmark of 60,000 handwritten digit images, we modified the feature distribution ($X$) in two distinct ways:

\textbf{Randomized Data Transformation}: We pre-processed the MNIST dataset (training and testing) by randomly selecting and transforming which inverts the colors (white-to-black and black-to-white) and maps to the different scales (mapping includes finding the maximum and minimum values, setting a rank, and then random a value r, and then process the scale mapping using \(((\text{max} - \text{min}) / \text{rank}) \cdot r + \text{min} + (x - \text{min}) / (\text{max} - \text{min})\)). The transformed data was distributed randomly among 100 clients, creating a mixture of multiple types of images per client. Figure \ref{fig:fed_mix} shows examples of client data. Then, we calculate the $H$ between each client data by using the Taylor approximate method. The mean of $H$ is $0.001468$ and the standard deviation is $0.001925$, which indicates the feature distributions between clients are similar.

\begin{figure}[H]
 \centering
 \begin{subfigure}[b]{0.49\linewidth}
 \includegraphics[width=\linewidth]{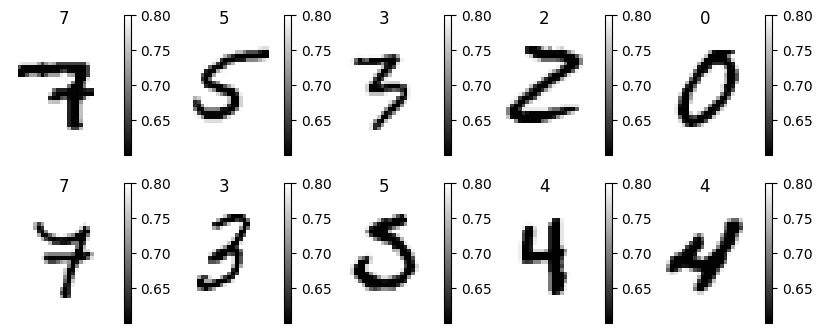} 
 \caption{Client (1)} 
 \end{subfigure}
 \hfill
 \begin{subfigure}[b]{0.49\linewidth}
 \includegraphics[width=\linewidth]{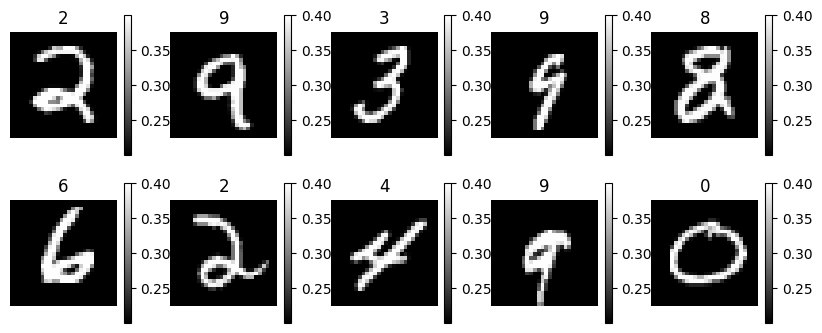} 
 \caption{Client (2)} 
 \end{subfigure}
 \caption{Different Feature Distribution MNIST Input }
 \label{fig:fed_diff} 
\end{figure}

\textbf{Feature-Based Allocation}: Data was distributed by type. For example, Client 1 exclusively received black numerals on white backgrounds and the values were within the specific scale, while Client 2 received white numerals on black backgrounds and the values were within another specific scale. Figure \ref{fig:fed_diff} illustrates these allocations. Then, we calculate the $H$ between each client data by using the Taylor approximate method. The mean of $H$ is $0.657$ and the standard deviation is $0.295$, which indicates the feature distributions between clients are different.

We evaluated the performance of a Multilayer Perceptron (MLP), optimized using Stochastic Gradient Descent (SGD) with a learning rate of 0.001. Training was conducted over 200 communication rounds with the Federated Averaging (FedAvg) framework.

\begin{figure}[H]
\centering
\includegraphics[width=0.4\textwidth]{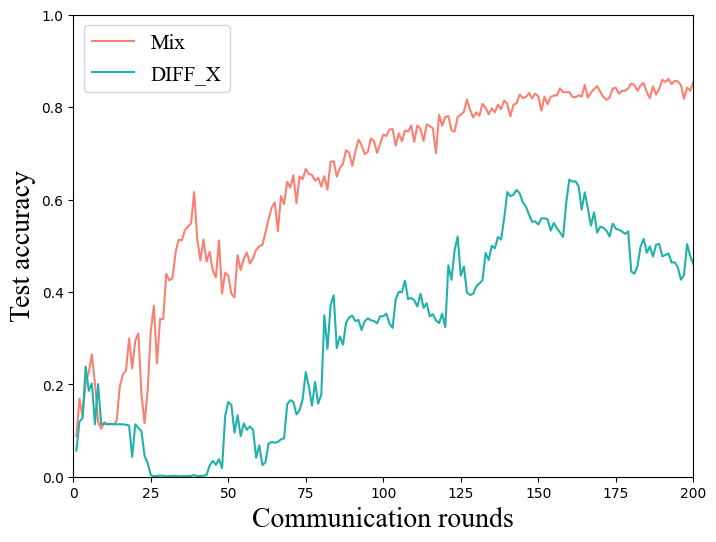} 
\caption{Test accuracy of MINST dataset for Federated Learning with different feature distribution.} 
\label{fig:fed_avg_result} 
\end{figure}

The experimental results revealed that models trained on mixed data outperformed those trained on distinct feature distributions, achieving higher accuracy and more stable convergence (Figure \ref{fig:fed_avg_result}). In contrast, the non-uniform availability of features in the distinct feature distribution setup led to significant declines in accuracy and stability during training. Additionally, analysis of the Energy Coefficient $H$ highlighted substantially greater heterogeneity in the treatment group, underscoring the sensitivity of federated learning models to feature distribution disparities among clients.

These findings underscore the importance of addressing feature heterogeneity in federated learning to ensure robust model performance and stability.


\section{Discussions and Conclusion}\label{sec:diss}


An important consideration in selecting a measure of feature heterogeneity is its ability to distinguish between distributions effectively. One alternative to the Energy Distance is a quadratic form of the distance, defined as

\begin{align*}
\tilde{D}^2(F_X, F_Y) = 2 \mathbb{E} \norm{X - Y}^2 - \mathbb{E} \norm{X - X'}^2 - \mathbb{E} \norm{Y - Y'}^2.
\end{align*}

While $\tilde{D}^2$ is non-negative and exhibits basic properties expected of a distance measure, it falls short in its ability to capture nuanced differences between distributions. Specifically, $\tilde{D}^2$ equals zero whenever $\mu_X = \mu_Y$ and $\sigma_X = \sigma_Y$, regardless of any disparities in higher-order moments. This limitation makes it ineffective for distinguishing distributions with identical means and variances but differing shapes or tails.

In contrast, the Energy Distance overcomes this limitation by accounting for differences in higher-order moments, in addition to location and scale. This property makes it a more suitable metric for quantifying feature heterogeneity in distributed scenarios, where comprehensive comparison of distributions is essential. While the Energy Distance may not address all potential limitations, its capacity to support hypothesis testing and capture broader distributional differences underscores its practical advantages over this quadratic alternative.

The Energy Coefficient $H$ provides a quantifiable measure of similarity between nodes and can be used to determine the penalty coefficient between guest and host models in distributed collaborative learning. By incorporating feature heterogeneity, $H$ enables dynamic adjustments that improve model alignment and overall system performance.

Feature heterogeneity extends to representation heterogeneity for categorical responses, enabling metrics to compute one distance per response value (e.g., a 2-dimensional vector for binary response). Challenges remain for continuous responses or cases with many unique discrete values.
Future work could refine the approximation by bounding residual terms or developing adaptive corrections for pronounced higher-order effects, enhancing accuracy and scalability.

In conclusion, we propose an efficient method to quantify heterogeneity in distributed learning. Simulations show its accuracy aligns with empirical results, offering a robust tool for analyzing feature distributions across clients and improving training strategies in distributed systems.

\bibliographystyle{chicago}
\bibliography{refer.bib}

\end{document}